\pdfoutput=1

\documentclass[11pt]{article}

\usepackage{acl}

\usepackage{times}
\usepackage{latexsym}

\usepackage[T1]{fontenc}

\usepackage[utf8]{inputenc}

\usepackage{graphicx}
\graphicspath{ {./images/} }
\usepackage{enumitem}
\usepackage{hyperref}
\usepackage{booktabs}
\usepackage{subcaption} 
\usepackage{longtable} 
\usepackage{array}

\title{Team Unibuc - NLP at SemEval-2024 Task 8: \\
Transformer and Hybrid Deep Learning Based Models for Machine-Generated Text Detection
}

\author{
Teodor-George Marchitan$^{1, 3, \thanks{\ \ Equal contributors}}$,
Claudiu Creanga$^{2, 3, \footnotemark[1]}$, Liviu P. Dinu$^{1, 3}$\\ 
  $^1$ Faculty of Mathematics and Computer Science, \\
  $^2$ Interdisciplinary School of Doctoral Studies, \\
  $^3$ HLT Research Center, \\
  University of Bucharest, Romania\\
  \small
{\tt teodor.marchitan@s.unibuc.ro, claudiu.creanga@s.unibuc.ro, ldinu@fmi.unibuc.ro}  \\
}

\vspace{2mm}
  
\begin{document}

\maketitle

\begin{abstract}

This paper describes the approach of the UniBuc - NLP team in tackling the SemEval 2024 Task 8: Multigenerator, Multidomain, and Multilingual Black-Box Machine-Generated Text Detection. We explored transformer-based and hybrid deep learning architectures. For subtask B, our transformer-based model achieved a strong \textbf{second-place} out of $77$ teams with an accuracy of \textbf{86.95\%}, demonstrating the architecture's suitability for this task. However, our models showed overfitting in subtask A which could potentially be fixed with less fine-tunning and increasing maximum sequence length. For subtask C (token-level classification), our hybrid model overfit during training, hindering its ability to detect transitions between human and machine-generated text.

\end{abstract}

\section{Introduction}
Task 8 from SemEval 2024 competition \cite{wang-EtAl:2024:SemEval20245} aims to tackle the complex challenge of distinguishing between human and AI generated text. Doing so is crucial for maintaining the integrity and authenticity of information as it helps prevent the spread of misinformation and ensures that content sources are accountable. By developing tools for this task, which work in a multilingual setting, and releasing them open source we can combat non-ethical uses of AI such as propaganda, misinformation, deepfakes, social manipulation and others. 

The systems developed for subtasks A and B are based on transformer models with different layers selection and merging strategies, followed by a set of fully connected layers. The training is split in two phases: a) freezing phase, where the transformer weights are not updated, only the fully connected layers are updated with a specific learning rate; b) fine-tuning phase, where the selected layers of the transformer and the fully connected layers are updated with a different (usually smaller) learning rate. For the subtask C, a different architecture was used, combining character level features, extracted with a CNN model, with word embeddings and fed into a Bidirectional LSTM followed by a set of fully connected layers. The same training strategy with different learning rates was used.

Our error analysis revealed that overfitting remains a primary challenge, despite our initial precautions. We learned that for future fine-tuning of transformer models, we should dedicate a lot more time to prevent overfitting. We made our models open source in a \href{https://github.com/ClaudiuCreanga/semeval-2024-task-8}{GitHub Repository}.

\section{Background}

\begin{figure}[htbp]
    \centering
    \includegraphics[width=0.9\linewidth]{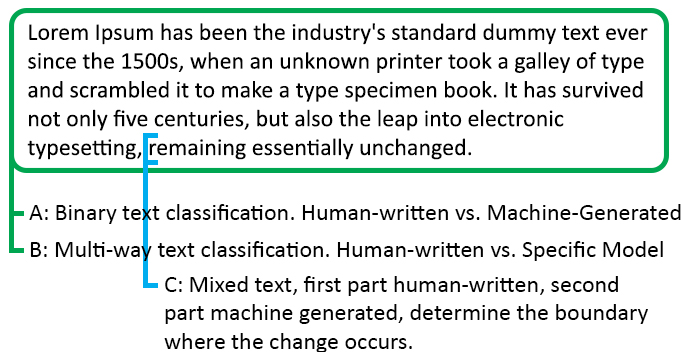}
    \caption{Three sub-tasks explained}
    \label{fig:three_subtasks}
\end{figure}

The competition had 3 tasks explained below (\autoref{fig:three_subtasks}). Subtask A had 2 sub-tracks: monolingual (English only) and multilingual. 

We participated in all 3 tasks with the best result being second place on subtask B (\autoref{table:team_results}).

\begin{table}
  \begin{tabular}{ccccc}
    \toprule
      & A mono & A multi & Track B & Track C \\
    \midrule
    \texttt \textbf{Score} & 85.13 & 79.43 & \textbf{86.95} & 74.28 \\
    \texttt \textbf{Place} & 33 / 137 & 30 / 68 & \textbf{2 / 77} & 31 / 33 \\
    \bottomrule
  \end{tabular}

  \caption{Team results}
  \label{table:team_results}
\end{table}

\subsection{Dataset}

The data for this task is an extension of the M4 dataset \cite{wang2024m4, wang2024mg-bench}. Compared to subtask A and B, subtask C had much less data to work with. We found out that we could increase the size of our datasets for subtask A monolingual by adding the dataset from subtask B and remove duplicated items (\autoref{table:dataset_sizes}). 
\begin{table}
  \begin{tabular}{ccccc}
    \toprule
      & A mono & A multi & Track B & Track C \\
    \midrule
    \texttt Train & 119757 & 172417 & 71027 & 3649 \\
    \texttt Dev & 5000 & 4000 & 3000 & 505 \\
    \texttt Test & 34272 & 42378 & 18000 & 11123 \\
    \bottomrule
  \end{tabular}

  \caption{Datasets sizes used in this competition by tasks.}
  \label{table:dataset_sizes}
\end{table}

\subsection{Previous Work}

Since GPT-2, it has been particularly difficult to detect machine-generated text, such that classical machine learning methods can no longer help. Previously, when models used top-k sampling, this resulted in text filled with too many common words and models could detect this anomaly easily \cite{ippolito-etal-2020-automatic}. But now with bigger and bigger models and other type of sampling (like nucleus sampling), fewer artifacts are left for a detector to spot. \citet{Solaiman2019ReleaseSA} showed that by fine-tuning a RoBERTa model we can achieve state of the art results for GPT-2 generated text with a 95\% accuracy. 

If for GPT-2, expert human evaluators achieved an accuracy of 70\% \cite{ippolito-etal-2020-automatic}, for GPT-3 and later models their accuracy is on par with random chance \cite{clark-etal-2021-thats}. It is still an open question if we can improve automated detection. Many companies (like OpenAI and Turnitin) are releasing products and claim to do it, but suffer from low rates of accuracy. \href{https://openai.com/blog/new-ai-classifier-for-indicating-ai-written-text}{In July 2023}, OpenAI removed its product for this reason.

\section{System overview}

In this paper, we focused our research on two different system architectures: \textbf{Transformer based models} (\ref{subsec:transformer_based}) and \textbf{Hybrid deep learning models} (\ref{subsec:hybrid_deep_learning_based}).

\begin{figure}[htbp]
    \centering
    \includegraphics[width=1\linewidth]{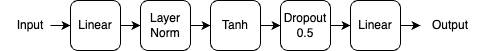}
    \caption{Fully connected layer base structure}
    \label{fig:fcc_layers}
\end{figure}

Both architectures use a block of fully connected layers (\autoref{fig:fcc_layers}) with the base structure being initiated with a linear layer, succeeded by normalization, a tanh activation function, followed by a dropout layer ($0.5$). Finally, it concludes with a linear layer with an output size of $1$ for subtask A and $6$ for subtask B .

\subsection{Transformer based models}\label{subsec:transformer_based}

\begin{figure}[htbp]
    \centering
    \includegraphics[width=0.8\linewidth]{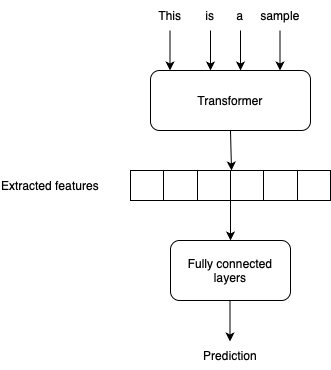}
    \caption{Transformer based models architecture}
    \label{fig:transformer_based_models_architecture}
\end{figure}

The core of this architecture is based on transformer models (\autoref{fig:transformer_based_models_architecture}). The strategy is to use the transformer model as a feature extractor, pass the information through fully connected layers (\autoref{fig:fcc_layers}) and apply the activation function based on the predictions for each task.

During the process of developing our system with this architecture, we encountered three difficulties that we had to address: 1) Long texts but limited number of tokens accepted by the transformer models (\ref{subsubsec:long_text_problem}); 2) Layer selection for feature extraction step (\ref{subsubsec:layer_selection}); 3) Fine-tuning strategy to prevent overfitting (\ref{subsubsec:fine_tuning_strategy}).

\subsubsection{Long text problem}\label{subsubsec:long_text_problem}

Most of the transformer models accept a maximum of $512$ tokens per sequence. We have also experimented the same strategies as described by \citet{sun2020finetune} in order to handle long texts.

\begin{enumerate}[label=\Roman*.]
    \item {
        \textbf{Truncation methods:}
        \begin{itemize}
            \item \label{itm:head_only} \textbf{Head only}: Keep only the first 510 tokens from the entire text. (extra 2 tokens for [CLS] and [SEP] tokens)
            \item \label{itm:tail_only} \textbf{Tail only}: Keep only the last 510 tokens from the entire text. (extra 2 tokens for [CLS] and [SEP] tokens)
            \item \label{itm:head_and_tail} \textbf{Head and Tail}: Combined the first $128$ tokens with the last $384$ tokens from the entire text.
        \end{itemize}
    }

    \item {
        \textbf{Hierarchical methods:} Each text is split into $k = L / 512$ chunks. For each chunk we get the pooled representation of [CLS] token and merge all chunk representations using mean or max.
    }
\end{enumerate}

Our experiments proved that truncation method with \textbf{head only} works best for the given dataset as well.

\subsubsection{Layer selection}\label{subsubsec:layer_selection}

Most transformer models have multiple layers and each layer is capturing different features from the input text \cite{sun2020finetune}. Intuitively, lower layers capture more general features at the token level and as we move up the layers, the captured features are more contextualized and more sensitive to the context of the tokens.

From our experiments, concatenating the last 4 layers and using only the last layer from the transformer proved to give the best results. Because of the limited resources, we chose to use only the last layer.

\subsubsection{Fine-tuning strategy}\label{subsubsec:fine_tuning_strategy}

Fine-tuning the transformer model for a downstream task is also challenging. Each layer of the transformer captures a different level of semantic and syntactic information from the input text \cite{yosinski2014transferable, howard2018universal, sun2020finetune}. We implemented a Head-First Fine-Tuning (HeFit) strategy \cite{Michail_2023} and used different learning rates for different layers \cite{sun2020finetune}:

\begin{enumerate}[label=\arabic*.]
    \item For the first number of epochs $[1,\ k]$  we completely freeze the transformer layers without updating any of the weights.
    \item For the rest of the epochs $[k + 1,\ N]$ we fine-tune only the selected layers used for feature extraction.
\end{enumerate}

Using this strategy, we are not only using less resources, but we can also preserve the more general information of the transformer (freezing lower layers) and updating information that is most relevant to the downstream task (fine-tuning selected upper layers).

\subsection{Hybrid deep learning models}\label{subsec:hybrid_deep_learning_based}

\begin{figure}[htbp]
    \centering
    \includegraphics[width=0.7\linewidth]{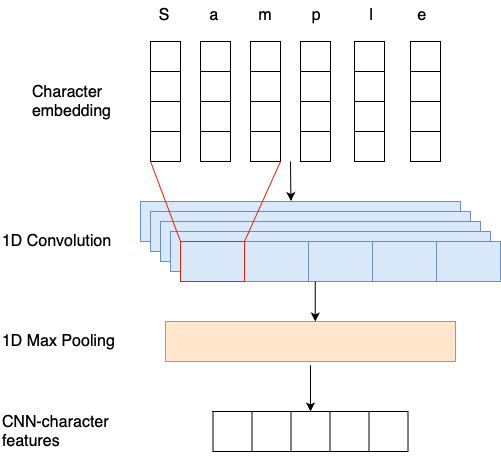}
    \caption{CNN-character level features}
    \label{fig:cnn_features}
\end{figure}

\begin{figure}[htbp]
    \centering
    \includegraphics[width=0.6\linewidth]{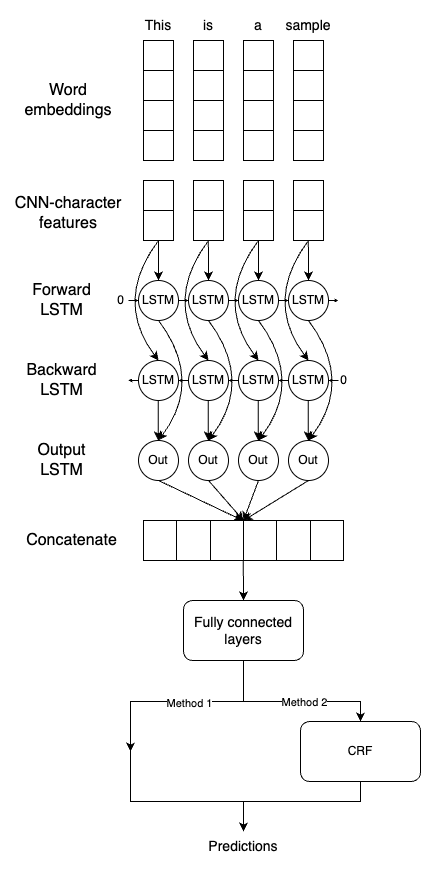} 
    \caption{Hybrid deep learning model architectures. Method 1 to use the predictions directly from the fully connected block and method 2 using CRF before predictions.}
    \label{fig:hybrid_deep_learning_models_architecture}
\end{figure}

This model architecture (\autoref{fig:hybrid_deep_learning_models_architecture}) was inspired by the work of \citet{chiu2016named} which proved to be very efficient for named entity recognition task. The idea was to convert words and characters into vector representations using lookup tables and concatenate them in order to be fed into a neural network. For the character-level features we used a lookup table for the character embeddings and applied a 1D convolution followed by a 1D max pooling layer (\autoref{fig:cnn_features}). For the word-level features we used a lookup table for the word embeddings. We concatenated the word and character features and fed them through a bidirectional LSTM and then a set of fully connected layers (\autoref{fig:hybrid_deep_learning_models_architecture} - method 1).

This model was mainly used for the subtask C, which we treated as a token classification task. Therefore we have also made some experiments adding a conditional random field \cite{sutton2010introduction} on top of the fully connected layers (\autoref{fig:hybrid_deep_learning_models_architecture} - method 2). This method was proved to be very efficient for sequence tagging by the work of \citet{huang2015bidirectional}.

\subsection{Experimental setup}
During the training phase, we utilized the development (dev) dataset as our test set, while the training dataset was divided into a training subset and a validation subset, following an 80\%-20\% split. For the construction of the final model, the entire training dataset was used for training purposes, with the dev set serving as our validation set. In terms of text preprocessing, we experimented with three different approaches:

\begin{itemize}
    \item \label{itm:Heavy} \textbf{Heavy}: Involved removing pre-trained language model special tokens such as <pad>, <s>, <unk>, etc., converting numbers into words, and eliminating special characters or formats like emails and URLs.
    \item \label{itm:Light} \textbf{Light}: Consisted of converting text to lowercase and removing special characters, including numbers.
    \item \label{itm:None} \textbf{None}: Text was used as is, without any preprocessing.
\end{itemize}

We observed that the model performed best with no preprocessing, a finding that aligns with the inherent flexibility of Masked Language Models to efficiently process raw text.

To determine the optimal number of training epochs, both when the pre-trained layers were kept frozen and during the fine-tuning phase, we monitored the validation set's loss and the test set's performance, opting for conservative epoch counts to prevent overfitting. 

\subsection{Subtask A}

For this subtask, in order to be able to run the models based on the transformer architectures, we used the head only truncation strategy (\ref{subsubsec:layer_selection} - \ref{itm:head_only}) with the first $512$ tokens.

\subsubsection{Monolingual}

In the monolingual track, the final submission is a transformer-based model architecture (\ref{subsec:transformer_based}) with RoBERTa-base pre-trained model. The extracted features from the transformer are only from the $[CLS]$ token of the last hidden layer with a $0.3$ dropout applied. The fully connected block is built with 2 base structures (\autoref{fig:fcc_layers}) consisting of $[256, 64]$ neurons. A $0.5$ dropout is applied and $sigmoid$ activation function is used in order to make the predictions. We trained this model in total for $5$ epochs with the entire transformer architecture freezed and a batch size of $24$ using the AdamW optimizer with a learning rate of $2e-4$ and the binary-cross entropy loss.

Regarding the layer selection, most of the experiments were done only using the last layer. We did some testing with last $4$ layers (for some pre-trained transformers) but we could not batch size $24$ anymore because of the limited resources if it were to also fine-tune the transformer's selected layers. We have also tested with multiple batch sizes and $24$ seemed to work best in our case. Results in \autoref{table:Task_A_experiments}.

\subsubsection{Multilingual}

For the multilingual track we used models pre-trained in a multilingual context (\autoref{table:Task_A_experiments_multilingual}) and for the final submission we chose mdeberta-v3-base which, even though it didn't support Indonesian, it gave the best results. The hyper-parameters that we chose were: batch size of $32$, token max length of $512$, a fully connecter layer (\autoref{fig:fcc_layers}) of $128$, learning rate for the "frozen step" of $0.001$ (where we train only the output layer) and smaller for fine-tuning: $0.0002$.

\subsection{Subtask B}

In the subtask B, the final submission is a transformer-based model architecture (\ref{subsec:transformer_based}) with RoBERTa-base pre-trained model. The extracted features from the transformer are only from the $[CLS]$ token of the last hidden layer with a $0.3$ dropout applied. The fully connected block is built with 2 base structures (\autoref{fig:fcc_layers}) consisting of $[512, 128]$ neurons and the final output size of the model being $6$. A $0.5$ dropout is applied with no activation function for making the predictions. We trained this model in total for $8$ epochs, first $6$ epochs with the entire transformer architecture freezed, and the last $2$ epochs also fine-tuning the last layer of the transformer (\ref{subsubsec:fine_tuning_strategy}). The batch size used was $32$ and optimizer AdamW with a learning rate of $3e-4$ for the freeze part of the training (updating only the fully-connected block weights) and $2e-4$ for the fine-tuning part with a linear scheduler with $50$ warmup steps and cross entropy loss.

Regarding the layer selection, most of the experiments were done only using the last layer. We did some testing with last $4$ layers (for some pre-trained transformers) but we could not batch size $32$ anymore because of the limited resources if it were to also fine-tune the transformer's selected layers. We have also tested with multiple batch sizes and $32$ seemed to work best in our case. Results in \autoref{table:Task_B_experiments}.

\subsection{Subtask C}

We treated this subtask as a token classification one and changed the labels from positions to list of $0$ and $1$, where $0$ means that the token at that specific position is not machine generated and $1$ otherwise. 

The tokenization was done by splitting the text by space and kept only the first $1024$ tokens from the entire text. 
As for the maximum character length of the tokens we went with $25$. 



The final submission is a hybrid deep learning model architecture (\ref{subsec:hybrid_deep_learning_based}). We used the method 2 variation of the architecture (\autoref{fig:hybrid_deep_learning_models_architecture} with the CRF model right before making the predictions.

For the CNN-character features we set the character embeddings dimension to $10$ and randomly initialized the lookup table using uniform distribution with range $[-0.5, 0.5]$. We used the convolution with kernel size $3$ and $20$ filters with a $0.5$ dropout afterwards. The word embedding dimension used is $300$ and the lookup table randomly initialized in the same manner. For the bidirectional LSTM we used $2$ filters with $32$ hidden dimension each. The fully connected block is build with a fully connected base structure (\ref{fig:fcc_layers}) with $32$ neurons and final output size of $2$.

We trained this model in total for $3$ epochs with a batch size of $12$ and optimizer AdamW with a learning rate of $5e-3$ for the first $2$ epochs of the training and $3e-3$ in the last epoch together  with a linear scheduler with no warmup steps and cross entropy loss.

\section{Results}

\subsection{Subtask A}
\begin{figure}[htbp]
    \centering
    \includegraphics[width=0.9\linewidth]{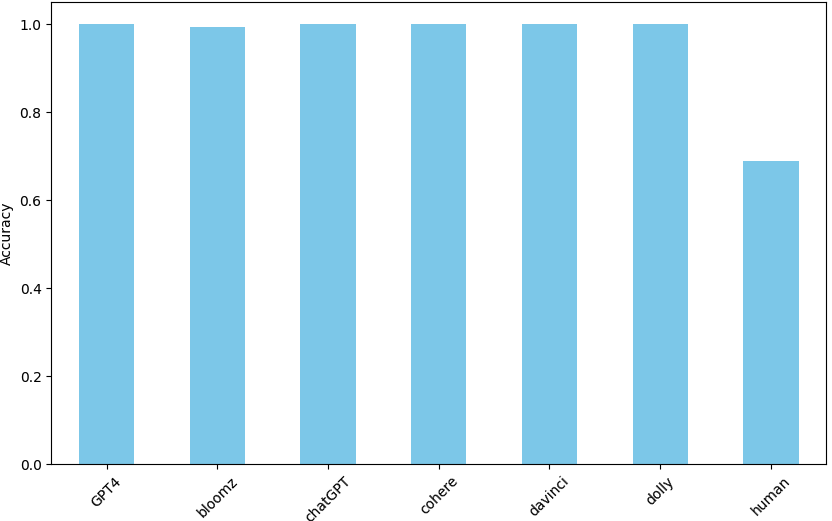}
    \caption{Subtask A: monolingual - accuracy by model for test set}
    \label{fig:monolingual_accuracy_by_model}
\end{figure}

For both monolingual and multilingual our model under-predicted the human-written class. In the case of the monolingual track our model performs equally well in detecting machine-generated text for each model, but under calls the negative class (\autoref{fig:monolingual_accuracy_by_model}). It predicts $23043$ items as machine generated and $11229$ as human-written while the truth was more balanced ($18000$ vs. $16272$). We obtain good accuracy for each machine generated model, but we under-call the human label (0.68 accuracy) so in the end the final score is 0.85.

\begin{figure}[htbp]
    \centering
    \includegraphics[width=0.9\linewidth]{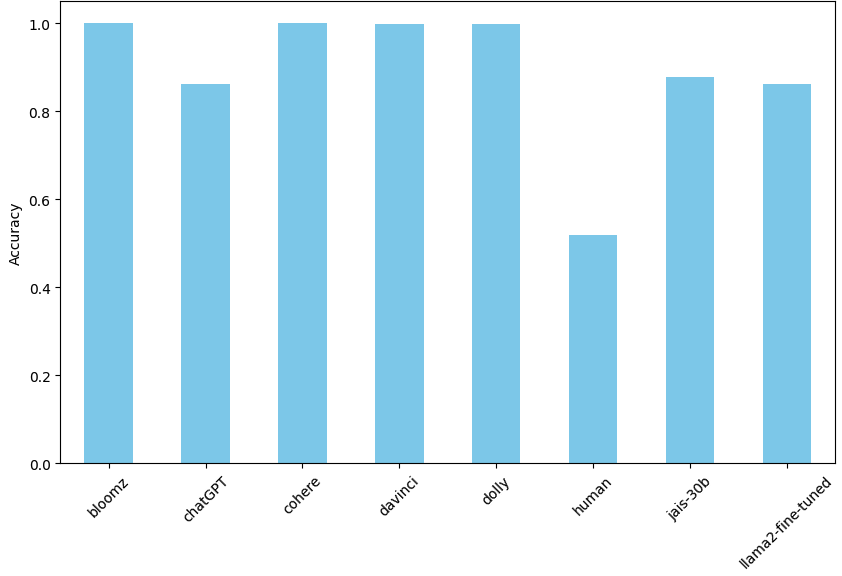}
    \caption{Subtask A: multilingual - accuracy by model for test set}
    \label{fig:multilingual_accuracy_by_model}
\end{figure}

In the case of multilingual, testing on dev data gave us an accuracy of $0.96$ but the final test score was $0.79$. Our model predicted $30764$ samples as machine generated and only 11614 as human-written, while the true distribution was more balanced ($22140$ vs. $20238$). This suggests that our model was overfit and had a bias for the positive class. If we look at the distribution per model we can see that we have a good accuracy on all models, except for human and a bit worse for chatGPT (\autoref{fig:multilingual_accuracy_by_model}), ending up with a final score of 0.79.

\begin{figure}[htbp]
    \centering
    \includegraphics[width=0.9\linewidth]{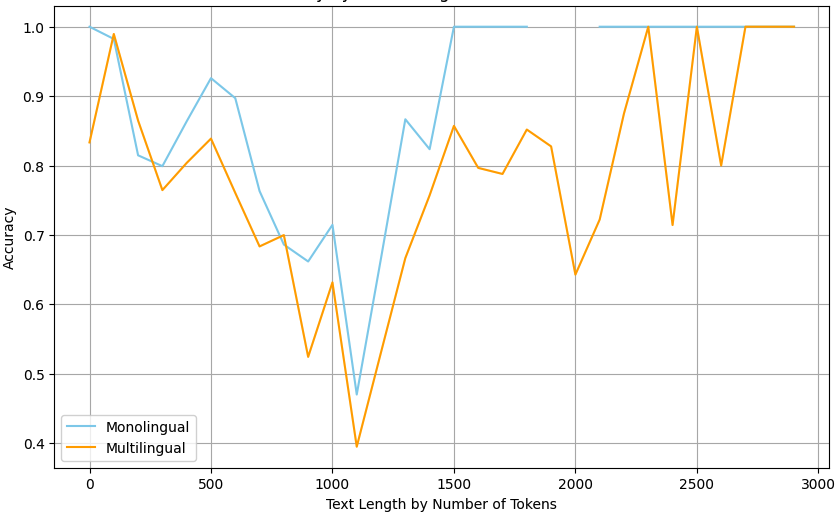}
    \caption{Subtask A: accuracy by sequence length in tokens, monolingual and multilingual}
    \label{fig:accuracy_by_token}
\end{figure}

If we look at sequence length we can see an U shaped graph at $500$ - $1500$ number of tokens, where the model performs worst (\autoref{fig:accuracy_by_token}) for both monolingual and multilingual tracks. We believe this is because our transformers had a limit of $512$ for token length and we didn't have the resources to train on a bigger sequence length. 

\subsection{Subtask B}

\begin{figure}[htbp]
    \centering
    \includegraphics[width=0.9\linewidth]{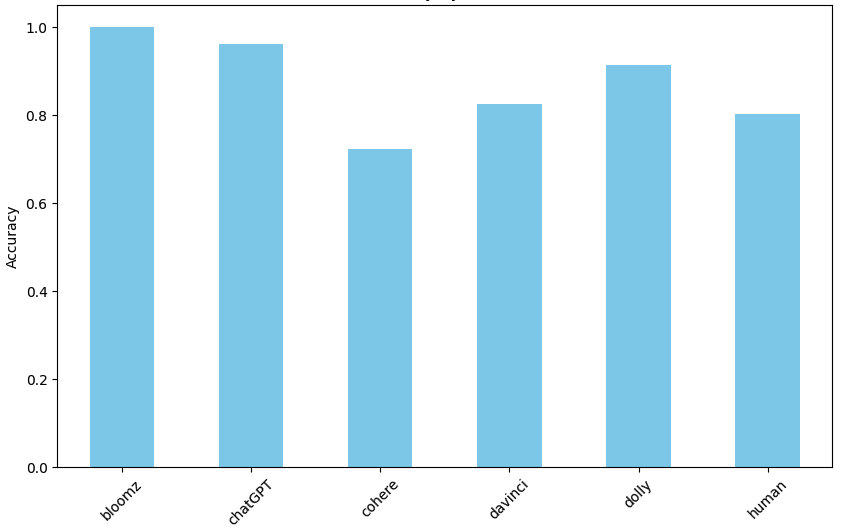}
    \caption{Subtask B: accuracy by model for test set.}
    \label{fig:subtask_B_accuracy}
\end{figure}

Our most notable performance was achieved in subtask B, where we secured the \textbf{second position} from a total of $77$ participating teams, with an accuracy score of \textbf{86.95\%}, very close to first position. Upon examining the accuracy breakdown by model, it becomes evident that our model exhibited strong performance, particularly with bloomz and chatGPT outputs, while facing more challenges with cohere (refer to \autoref{fig:subtask_B_accuracy}). The elevated score compared to Task A implies that our model's architecture and training methodology were well-suited for the demands of a multiclass classification task.

\subsection{Subtask C}

Our results on the subtask C show that the model architecture we chose alongside the hyperparameters overfitted drastically on this dataset. The MAE on training data decreased from $18.8$ to $4.39$ and on validation data decreased from $18.04$ to $8.34$ 
during the training phase, while on the final test dataset the MAE increased to $74.28$. This proves that the character and word embeddings could not generalize that good in order to be able to find that transition spot from human text to machine generated text.


\section{Conclusions and Future Work}

In conclusion, our architecture and training methods produced good results for subtask B (securing the second place). However, our models demonstrated signs of overfitting for subtask A. We could not find a proper explanation for why the model architecture work better on subtask B and is overfitting that much on the other task. Our future endeavors will explore several avenues:
\begin{itemize}
    \item \textbf{Extended Sequence Lengths}: With more powerful machines we plan to increase the token length from 512 to 1024 in order to capture a wider context, which could improve their performance.
    \item \textbf{Ensemble Learning with Model Specialization}: Split the dataset by originating model (chatGPT, cohere etc.) and train specialized models on each subset. Each specialized model will become adept at discerning text generated by its corresponding model. By aggregating predictions from these specialized models, we aim to construct a meta-model capable of making better final predictions.
    \item \textbf{LLM}: We plan to investigate the efficacy of large language models (like Mistral/Mixtral or Solar) with either zero shot learning or few shot learning scenarios. For few-shot learning, we intend to exploit the in-context learning capabilities of LLMs by presenting them with pairs of examples (one human-written and one machine-generated) within the same context window. We will then ask the model to predict an unseen example.
\end{itemize}

\section*{Acknowledgements}

This work was partially supported by a grant on Machine Reading Comprehension from Accenture Labs and by the POCIDIF project in Action 1.2. "Romanian Hub for Artificial Intelligence".

\bibliography{anthology}

\begin{thebibliography}{13}
\expandafter\ifx\csname natexlab\endcsname\relax\def\natexlab#1{#1}\fi

\bibitem[{Chiu and Nichols(2016)}]{chiu2016named}
Jason P.~C. Chiu and Eric Nichols. 2016.
\newblock \href {http://arxiv.org/abs/1511.08308} {Named entity recognition with bidirectional lstm-cnns}.

\bibitem[{Clark et~al.(2021)Clark, August, Serrano, Haduong, Gururangan, and Smith}]{clark-etal-2021-thats}
Elizabeth Clark, Tal August, Sofia Serrano, Nikita Haduong, Suchin Gururangan, and Noah~A. Smith. 2021.
\newblock \href {https://doi.org/10.18653/v1/2021.acl-long.565} {All that{'}s {`}human{'} is not gold: Evaluating human evaluation of generated text}.
\newblock In \emph{Proceedings of the 59th Annual Meeting of the Association for Computational Linguistics and the 11th International Joint Conference on Natural Language Processing (Volume 1: Long Papers)}, pages 7282--7296, Online. Association for Computational Linguistics.

\bibitem[{Howard and Ruder(2018)}]{howard2018universal}
Jeremy Howard and Sebastian Ruder. 2018.
\newblock \href {http://arxiv.org/abs/1801.06146} {Universal language model fine-tuning for text classification}.

\bibitem[{Huang et~al.(2015)Huang, Xu, and Yu}]{huang2015bidirectional}
Zhiheng Huang, Wei Xu, and Kai Yu. 2015.
\newblock \href {http://arxiv.org/abs/1508.01991} {Bidirectional lstm-crf models for sequence tagging}.

\bibitem[{Ippolito et~al.(2020)Ippolito, Duckworth, Callison-Burch, and Eck}]{ippolito-etal-2020-automatic}
Daphne Ippolito, Daniel Duckworth, Chris Callison-Burch, and Douglas Eck. 2020.
\newblock \href {https://doi.org/10.18653/v1/2020.acl-main.164} {Automatic detection of generated text is easiest when humans are fooled}.
\newblock In \emph{Proceedings of the 58th Annual Meeting of the Association for Computational Linguistics}, pages 1808--1822, Online. Association for Computational Linguistics.

\bibitem[{Michail et~al.(2023)Michail, Konstantinou, and Clematide}]{Michail_2023}
Andrianos Michail, Stefanos Konstantinou, and Simon Clematide. 2023.
\newblock \href {https://doi.org/10.18653/v1/2023.semeval-1.140} {Uzh clyp at semeval-2023 task 9: Head-first fine-tuning and chatgpt data generation for cross-lingual learning in tweet intimacy prediction}.
\newblock In \emph{Proceedings of the The 17th International Workshop on Semantic Evaluation (SemEval-2023)}. Association for Computational Linguistics.

\bibitem[{Solaiman et~al.(2019)Solaiman, Brundage, Clark, Askell, Herbert-Voss, Wu, Radford, and Wang}]{Solaiman2019ReleaseSA}
Irene Solaiman, Miles Brundage, Jack Clark, Amanda Askell, Ariel Herbert-Voss, Jeff Wu, Alec Radford, and Jasmine Wang. 2019.
\newblock \href {https://api.semanticscholar.org/CorpusID:201666234} {Release strategies and the social impacts of language models}.
\newblock \emph{ArXiv}, abs/1908.09203.

\bibitem[{Sun et~al.(2020)Sun, Qiu, Xu, and Huang}]{sun2020finetune}
Chi Sun, Xipeng Qiu, Yige Xu, and Xuanjing Huang. 2020.
\newblock \href {http://arxiv.org/abs/1905.05583} {How to fine-tune bert for text classification?}

\bibitem[{Sutton and McCallum(2010)}]{sutton2010introduction}
Charles Sutton and Andrew McCallum. 2010.
\newblock \href {http://arxiv.org/abs/1011.4088} {An introduction to conditional random fields}.

\bibitem[{Wang et~al.(2024{\natexlab{a}})Wang, Mansurov, Ivanov, su, Shelmanov, Tsvigun, Mohammed~Afzal, Mahmoud, Puccetti, Arnold, Whitehouse, Aji, Habash, Gurevych, and Nakov}]{wang-EtAl:2024:SemEval20245}
Yuxia Wang, Jonibek Mansurov, Petar Ivanov, jinyan su, Artem Shelmanov, Akim Tsvigun, Osama Mohammed~Afzal, Tarek Mahmoud, Giovanni Puccetti, Thomas Arnold, Chenxi Whitehouse, Alham~Fikri Aji, Nizar Habash, Iryna Gurevych, and Preslav Nakov. 2024{\natexlab{a}}.
\newblock \href {https://aclanthology.org/2024.semeval2024-1.275} {Semeval-2024 task 8: Multidomain, multimodel and multilingual machine-generated text detection}.
\newblock In \emph{Proceedings of the 18th International Workshop on Semantic Evaluation (SemEval-2024)}, pages 2041--2063, Mexico City, Mexico. Association for Computational Linguistics.

\bibitem[{Wang et~al.(2024{\natexlab{b}})Wang, Mansurov, Ivanov, Su, Shelmanov, Tsvigun, Whitehouse, Afzal, Mahmoud, Sasaki, Arnold, Aji, Habash, Gurevych, and Nakov}]{wang2024m4}
Yuxia Wang, Jonibek Mansurov, Petar Ivanov, Jinyan Su, Artem Shelmanov, Akim Tsvigun, Chenxi Whitehouse, Osama~Mohammed Afzal, Tarek Mahmoud, Toru Sasaki, Thomas Arnold, Alham~Fikri Aji, Nizar Habash, Iryna Gurevych, and Preslav Nakov. 2024{\natexlab{b}}.
\newblock M4: Multi-generator, multi-domain, and multi-lingual black-box machine-generated text detection.
\newblock In \emph{Proceedings of the 18th Conference of the European Chapter of the Association for Computational Linguistics}, Malta.

\bibitem[{Wang et~al.(2024{\natexlab{c}})Wang, Mansurov, Ivanov, Tsvigun, Su, Shelmanov, Afzal, Mahmoud, Puccetti, Arnold, Aji, Habash, Gurevych, and Nakov}]{wang2024mg-bench}
Yuxia Wang, Jonibek Mansurov, Petar Ivanov, Akim Tsvigun, Jinyan Su, Artem Shelmanov, Osama~Mohammed Afzal, Tarek Mahmoud, Giovanni Puccetti, Thomas Arnold, Alham~Fikri Aji, Nizar Habash, Iryna Gurevych, and Preslav Nakov. 2024{\natexlab{c}}.
\newblock {MG-Bench}: Evaluation benchmark for black-box machine-generated text detection.

\bibitem[{Yosinski et~al.(2014)Yosinski, Clune, Bengio, and Lipson}]{yosinski2014transferable}
Jason Yosinski, Jeff Clune, Yoshua Bengio, and Hod Lipson. 2014.
\newblock \href {http://arxiv.org/abs/1411.1792} {How transferable are features in deep neural networks?}

\end{thebibliography}

\appendix

\section{Further experiments - Subtask A}

For most of the experiments in subtask A monolingual, we used two fully connected layers (\ref{fig:fcc_layers}) with $[256, 64]$ neurons, batch size $24$ and trained the model in total for $5$ epochs. For all experiments we used AdamW optimizer with learning rate $2e-4$ and binary-cross entropy loss. For some of the experiments we have also tried fine-tuning the last $n$ selected layers (in most cases just the last layer) for the last $k$ epochs. In those cases, we have also used a linear scheduler with $50$ warmup steps and changed the learning rate as well. The results can be seen in \autoref{table:Task_A_experiments}. Experiments for the multilingual track kept the same architecture as the monolingual one but used multilingual pre-trained models \autoref{table:Task_A_experiments_multilingual}. 

\begin{table}[htbp]
    \centering
    \resizebox{0.4\textwidth}{!}{%
    \begin{tabular}{ccccc}
        \toprule
        Model & Train & Validation & Test & Final \\
        \midrule
        \texttt mdeberta-v3 & 0.96 & 0.95 & 0.94 & \textbf{0.79} \\
        \texttt xlm-roberta & 0.97 & 0.95 & 0.92 & 0.78 \\
        \texttt bert-multi & 0.95 & 0.92 & 0.91 & 0.75 \\
        \texttt distilbert-multi & 0.93 & 0.90 & 0.89 & 0.73 \\
        \bottomrule
    \end{tabular}
    }
    \caption{Experiment results by pre-trained model - multilingual. Validation was the dev set, test size was 0.2 and final score is the test score in competition.}
    \label{table:Task_A_experiments_multilingual}
\end{table}

\section{Further experiments - Subtask B}

For most of the experiments in subtask B, we used two fully connected layers (\ref{fig:fcc_layers}) with $[512, 128]$ neurons, batch size $32$ and a trained the model in total $8$ epochs. For all experiments we used AdamW optimizer with learning rate $3e-4$ and cross entropy loss. For some of the experiments we have also tried fine-tuning the last $n$ selected layers (in most cases just the last layer) for the last $k$ epochs. In those cases, we have also used a linear scheduler with $50$ warmup steps and changed the learning rate as well. The results can be seen in \autoref{table:Task_B_experiments}.

\begin{table*}[bt]
\begin{center}
    \resizebox{0.9\textwidth}{!}{%
    \begin{tabular}{ccccccc}
        \toprule
        Base model & Epochs before fine-tune & LR fine-tune & Train & Validation & Test & Final \\
        \midrule
        \texttt roberta-base & 5 & --- & 0.89 & 0.94 & 0.89 & \textbf{0.85} \\
        \texttt flan-t5-base & 5 & --- & 0.98 & 0.97 & 0.95 & 0.84 \\
        \texttt deberta-v3-large & 5 & --- & 0.98 & 0.97 & 0.96 & \textbf{0.85} \\
        \texttt albert-base-v2 & 5 & --- & 0.77 & 0.82 & 0.74 & 0.83 \\
        \texttt bert-base-cased & 5 & --- & 0.79 & 0.80 & 0.76 & \textbf{0.86} \\
        \texttt distilbert-base-uncased & 5 & --- & 0.84 & 0.85 & 0.79 & 0.74 \\
        \texttt gpt2 & 5 & --- & 0.92 & 0.92 & 0.86 & 0.76 \\
        \texttt xlm-roberta-base & 5 & --- & 0.74 & 0.79 & 0.75 & 0.83 \\
        \texttt xlnet-base-cased & 5 & --- & 0.74 & 0.80 & 0.79 & 0.79 \\
        \texttt roberta-base & 4 & 0.0002 & 0.88 & 0.92 & 0.88 & 0.83 \\
        \texttt roberta-base & 3 & 0.0001 & 0.99 & 0.99 & 0.93 & 0.68 \\
        \bottomrule
    \end{tabular}
    }

    \caption{Experiment results for Subtask A - monolingual track. Validation was the dev set, test size was 0.2 and final score is the test score in competition.}
    
    \label{table:Task_A_experiments}
\end{center}


\bigskip

  



\begin{minipage}{\textwidth}
\centering

\resizebox{0.9\textwidth}{!}{%
\begin{tabular}{cccccccc}
    \toprule
    Base model & Epochs & Epochs before fine-tune & LR fine-tune & Train & Validation & Test & Final \\
    \midrule
    \texttt roberta-base & 8 & 6 & 0.0002 & 0.98 & 0.97 & 0.90 & \textbf{0.87} \\
    \texttt roberta-base & 6 & 6 & --- & 0.76 & 0.86 & 0.74 & 0.59 \\
    \texttt bert-base-cased & 8 &  6 & 0.0002 & 0.92 & 0.88 & 0.90 & 0.57 \\
    \texttt bert-base-cased & 6 & 6 & --- & 0.67 & 0.76 & 0.63 & 0.47 \\
    \bottomrule
\end{tabular}
}

\caption{Experiment results for Subtask B. Validation was the dev set, test size was 0.2 and final score is the test score in competition.}

\label{table:Task_B_experiments}

\end{minipage}

\end{table*}



\end{document}